\PassOptionsToPackage{table}{xcolor}
\pdfoutput=1
\documentclass[11pt]{article}
\usepackage[preprint]{acl}
\usepackage{times}
\usepackage{latexsym}
\usepackage[T1]{fontenc}
\usepackage[utf8]{inputenc}
\usepackage{microtype}
\usepackage{inconsolata}
\usepackage{amsmath}
\usepackage{booktabs}
\usepackage{multirow}
\usepackage{amssymb}
\usepackage{balance}
\usepackage{subcaption}

\usepackage{pifont}
\usepackage{makecell}
\usepackage{filecontents}
\usepackage{bbm}
\usepackage{pgfplots}
\usetikzlibrary{patterns}
\usepackage[inline,shortlabels]{enumitem}
\usepackage{natbib}
\usepackage{csquotes}
\usepackage{hyperref}
\usepackage{url}
\usepackage{CJKutf8}
\usepackage{algorithm, algorithmicx, algpseudocode}
\usepackage{svg}
\usepackage{bm}
\usepackage{graphics}
\usepackage{graphicx}
\usepackage{array}
\usepackage[english]{babel}
\usepackage{textcomp}
\usepackage{latexsym}
\usepackage{siunitx}  
\usepackage[normalem]{ulem}
\useunder{\uline}{\ul}{}
\usepackage{array}
\usepackage{titletoc}
\usepackage{tikz}
\usepackage{longtable}
\usetikzlibrary{tikzmark}
\usepackage[most]{tcolorbox}
\makeatletter
\newcommand*\myfontsize{%
  \@setfontsize\myfontsize{7}{8}%
}
\makeatother
\newcommand{\mytextbox}[2]{\tikzmarknode[draw=#1,thick,inner sep=2pt,outer sep=2pt]{test}{\myfontsize #2}}

\definecolor{myred}{rgb}{0.7, 0.3, 0.0}
\definecolor{myblue}{rgb}{0.2, 0.3, 0.6}
\definecolor{mygreen}{HTML}{008000}

\newcommand{\ours}{LongRefiner}

\newcommand{\mysection}{\mytextbox{myblue}{\textbf{\textcolor{myblue}{<section: \{title\}>}}}}

\newcommand{\mysubsection}{\mytextbox{myblue}{\textbf{\textcolor{myblue}{<subsection: \{title\}>}}}}
\newcommand{\mysubsectionend}{\mytextbox{myblue}{\textbf{\textcolor{myblue}{</subsection: \{title\}>}}}}
\newcommand{\myskip}{\mytextbox{myred}{\textbf{\textcolor{myred}{<skip>}}}}
\newcommand{\mybr}{\mytextbox{myred}{\textbf{\textcolor{myred}{<br>}}}}

\newcommand{\myglobal}{\mytextbox{mygreen}{\textbf{\textcolor{mygreen}{Global}}}}
\newcommand{\mylocal}{\mytextbox{mygreen}{\textbf{\textcolor{mygreen}{Local}}}}

\title{Hierarchical Document Refinement for Long-context \\
Retrieval-augmented Generation}

\author{%
Jiajie Jin$^1$, Xiaoxi Li$^1$, Guanting Dong$^1$, Yuyao Zhang$^1$ \\
\textbf{Yutao Zhu$^1$,} \textbf{Yongkang Wu$^2$,} \textbf{Zhonghua Li$^2$,}
\textbf{Qi Ye$^2$,} \textbf{Zhicheng Dou$^1$}\thanks{Correpsonding author.} \\
$^1$Gaoling School of Artificial Intelligence, Renmin University of China\\
$^2$Huawei Poisson Lab\\
\texttt{\{jinjiajie, dou\}@ruc.edu.cn} \\
}

\begin{document}
\begin{CJK}{UTF8}{gbsn}

\maketitle

\begin{abstract}

Real-world RAG applications often encounter long-context input scenarios, where redundant information and noise results in higher inference costs and reduced performance. 
To address these challenges, we propose \textbf{\ours{}}, an efficient plug-and-play refiner that leverages the inherent structural characteristics of long documents. \ours{} employs dual-level query analysis, hierarchical document structuring, and adaptive refinement through multi-task learning on a single foundation model. 
Experiments on seven QA datasets demonstrate that \ours{} achieves competitive performance in various scenarios while using 10x fewer computational costs and latency compared to the best baseline.
Further analysis validates that \ours{} is scalable, efficient, and effective, providing practical insights for real-world long-text RAG applications. Our code is available at \url{https://github.com/ignorejjj/LongRefiner}.


\end{abstract}

\section{Introduction}
\label{sec1: intro}

Large language models (LLMs) have demonstrated remarkable capabilities and achieved impressive results in various applications~\cite{llm-survey,survey_hallu_llm}.
However, due to their capabilities are limited to the training data, they are unable to update their knowledge in real-time~\cite{searcho1}, leading to poor performance in knowledge-intensive tasks~\cite{kilt} and factual accuracy~\cite{survey-factuality, Li2025WebThinker}. Retrieval-augmented generation (RAG)~\cite{rag,retro} addresses these limitations by combining information retrieval techniques with generative models, enabling access to external knowledge bases and significantly improving the accuracy and reliability of generated content~\cite{ trustworthy_survey}.

While the effectiveness of RAG systems critically depends on the quality and information density of retrieved content~\cite{survey_hallu_llm,ragsurvey,bider_jiajie}, real-world scenarios present significant challenges when dealing with lengthy documents returned by retrievers such as search engines~\cite{24_acl_chunking_free, longllmlingua}. Although these documents contain the necessary information for generating accurate responses, their extensive length poses two primary challenges for practical RAG deployments: (1)~\textbf{Signal-to-noise ratio:} Long documents often contain substantial irrelevant content alongside pertinent information, making it difficult for models to focus on query-relevant details~\cite{promptcompress_survey,genir_survey, bider_jiajie}. (2)~\textbf{Computational overhead:} Processing complete documents significantly increases the input context length, resulting in higher computational costs and potential performance bottlenecks in production environments~\cite{llm-survey}.

\begin{figure}[!t]
\centering
\includegraphics[width=0.8\linewidth]{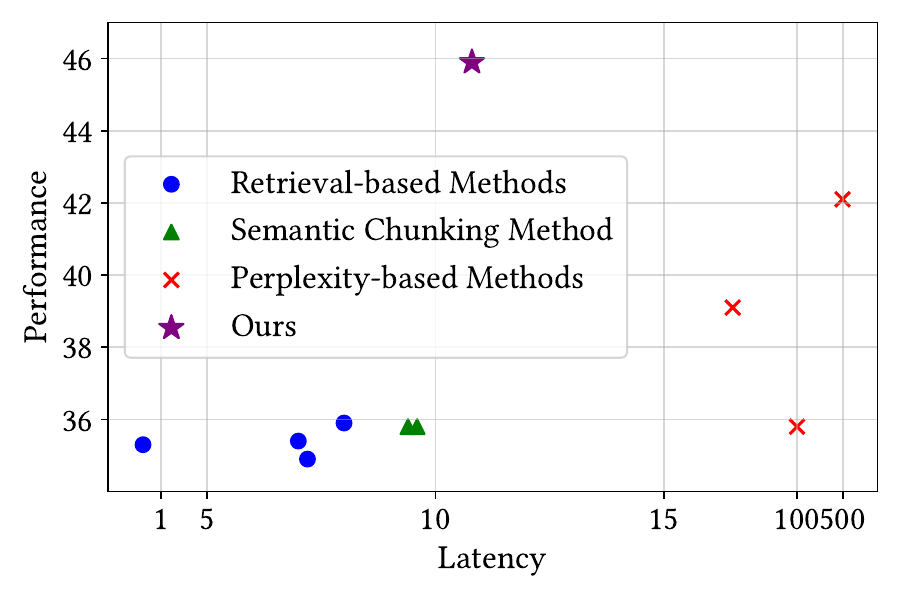}
\caption{Comparison of different methods in terms of efficiency and effectiveness.}
\label{fig:empirical_study}
\end{figure}

To address these challenges, an intuitive approach is to refine long retrieved documents before LLM processing~\cite{longllmlingua}. Unfortunately, current refinement methods typically either suitable for short text chunks or rely on crude metrics like perplexity to assess token relevance~\cite{selective-context}. 
As shown in Figure~\ref{fig:empirical_study}, these approaches fail to effectively utilize complete document information due to limited query understanding and global context awareness, leading to performance degradation and high latency. However, we observe that complete documents contain rich structural information like logical connection and content organization, which can enable more precise information extraction than traditional chunk-based approaches.

Motivated by these observations, we aim to achieve efficient document refinement by modeling structural information in long documents. To this end, we propose \textbf{\ours{}}, a plug-and-play efficient refinement system for long retrieved documents. \ours{} integrates three key capabilities: dual-level query analysis, hierarchical document structuring, and adaptive refinement, combining them through multi-task LoRA~\cite{lora} learning on a single foundation model to enhance overall usability. To improve system efficiency, we design a simplified XML-based syntax~\cite{xmlviews,common_html} for representing document structure, which significantly reduces the refinement model's output token count. Furthermore, we developed an efficient inference paradigm that achieves low online latency by executing certain tasks offline.
Our experiments across seven diverse QA datasets demonstrate superior performance over existing baselines across various query types while maintaining lower latency. We further validate practical feasibility through extensive experiments with different backbone models and training data scales. 

Our main contributions are:
(1) We propose a universal document-level refinement framework that achieves efficient, low-latency long-text refinement by leveraging hierarchical textual information.
(2) To address the challenges of noise and low information density in long retrieved documents, \ours{} introduces three key steps: dual-level query analysis, hierarchical document structuring, and adaptive document refinement, significantly optimizing RAG costs and response latency.
(3) We develop an efficient training and inference paradigm, achieving low online latency through LoRA-based multi-task learning combined with offline and online task orchestration.
(4) Experimental results demonstrate that our approach achieves superior generation quality with only 10\% of the token budget compared to existing text compression methods, while maintaining lower latency.

\section{Problem Formulation}
In a standard RAG pipeline, the retriever retrieves relevant documents from corpus $\mathcal{C}$ based on a query $\mathcal{Q}$, and the system then constructs the input prompt by combining the retrieved documents, query, and instruction $\mathcal{I}$. This prompt is input to the generation model to obtain the response. To enhance the signal-to-noise ratio, we introduce a refiner module to distill the retrieved documents.

Given a fixed retriever $\mathcal{R}$, a corpus $\mathcal{C}$, and a generator $\mathcal{G}$, with each query yielding a set of retrieved documents $D$, we seek a mapping function $\mathcal{F}$ that transforms retrieved documents into refined content. The effectiveness of $\mathcal{F}$ can be measured through: (1) downstream performance, measured by $\mathcal{G}(\mathcal{A}|\mathcal{F}(D))$, where $\mathcal{A}$ is the golden answer; (2) compression ratio $\gamma = |D|/|\mathcal{F}(D)|$, defined as the token count ratio before and after mapping; and (3) computational latency $\tau$, which is the execution time of the mapping function itself. Our goal is to design an efficient mapping function that optimizes downstream performance while minimizing latency under a fixed compression ratio.

\section{\ours{}: a Long Document Refiner for RAG}

\begin{figure*}[!t]
    \centering
    \includegraphics[width=0.85\linewidth]{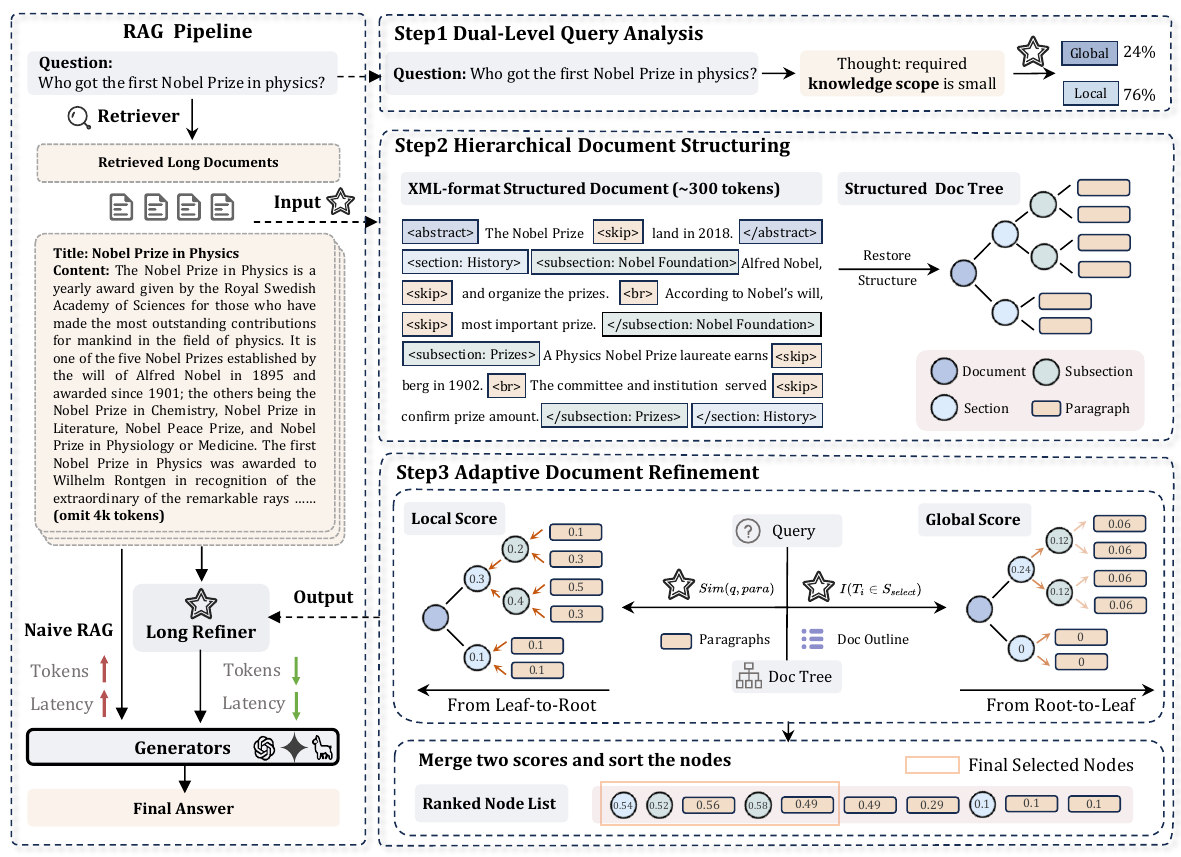}
    \caption{
    Overview of the \ours{} Framework.
    }
    \label{fig:overview}
\end{figure*}

To address the challenges of long-context RAG outlined in Introduction, our approach focuses on designing an efficient, query-aware long document refiner based on hierarchical modeling of long texts. In this section, we present three key steps of our method, followed by a comprehensive description of our training and inference process.

\subsection{Dual-Level Query Analysis}

In real-world scenarios, queries exhibit diverse information needs ranging from simple facts to complex reasoning~\cite{slimplm_jiejun,rqrag}. To characterize this diversity, we introduce two information levels: \textbf{Local Level} and \textbf{Global Level}. The Local Level refers to knowledge confined to specific contexts or localized information, involving a narrow knowledge scope such as a single passage or snippet. In contrast, the Global Level encompasses broad, comprehensive knowledge, involving a wide range of information and background context. 

To quantify these levels, we construct a dual-level query analyzer. Specifically, we first prompt a teacher LLM with task-specific instructions to analyze each query in the training dataset and assign a corresponding information level, which is represented as a binary label.\footnote{Instructions are shown in Appendix~\ref{appendix:instructions}} We then treat this label as a special token and finetune the refiner to generate the corresponding special token based on the input query. During inference, \ours{} adaptively determines the amount of information required for each query by predicting the appropriate information-level token. The formula can be written as:
\begin{equation}
\begin{aligned}
\label{equ:query_ana}
    &P_l = P_\mathcal{M}(\mylocal \mid  \text{query}) \\
    &P_g = P_\mathcal{M}(\myglobal \mid \text{query})  \\
    &R_q = \text{Softmax}(P_l, P_g)_g. 
\end{aligned}
\end{equation}

As shown in Equation~\ref{equ:query_ana}, to achieve a more precise representation of the information scope rather than a simple binary label, we apply a $\operatorname{softmax}$ operation to the generation probabilities of the two level tokens, producing a continuous representation $r_q$ as the information scope for the query.

\subsection{Hierarchical Document Structuring}

To facilitate efficient refinement, we convert unstructured retrieved documents into a hierarchical format with clear article outline, hierarchical organization, and paragraph segmentation. The hierarchical structure is defined as follows.
\paragraph{Tree-based structured document definition.}  
We model the structured document as a doc tree. Formally, the structured representation of a document \(D\) is denoted as \(D_{\text{str}} = (\mathcal{N}, \mathcal{R})\), where \(\mathcal{N}\) is the set of nodes in the document. As shown in the structured doc tree in Figure~\ref{fig:overview}, each node represents a section, subsection, or paragraph, with its corresponding title and content. \(R\) denotes the set of relations, which capture the implication and hierarchical relationships between nodes. 

To derive such structured representation from the original long document, we leverage a long-context window LLM as our backbone. However, this task still poses two significant challenges: (1) As \(D_{\text{str}}\) introduces additional information such as titles and structural information, the number of tokens required to represent it is greater than the tokens in \(D\) itself, which results in a excessively long learning target. (2) The retrieved documents consist only of plain text without the full structure or outline, lacking the golden labels needed for training.
To address these challenges, we propose two novel designs.
\paragraph{XML-format document structure representation.}  

\begin{table}[!t]
\small 
\centering
\caption{XML-format tags and definitions. \ours{} will use these tags to represent the structural information of the document during the generation process.}
\begin{tabular}{@{}ll@{}}
\toprule
\textbf{Tag Format} & \textbf{Definitions} \\
\midrule
\mysection & Begin of the section with specify title \\
\mysubsection & Begin of the subsection with specify title \\
\myskip & Placeholder for omitting middle content \\
\mybr & Paragraph switching symbol \\
\bottomrule
\end{tabular}
\label{tab:tokens}
\end{table}

Inspired by XML syntax and web page representation, we design an XML-format doc structuring method to address the problem of document being too long and difficult to learn.
We first design a flat representation \(D_{\text{xml}}\) of \(D_{\text{str}}\), which corresponds to \(D_{\text{str}}\) one-to-one, enabling the complete representation of the document's overall structure with fewer tokens.

An example of $D_{\text{str}}$ is shown in Figure~\ref{fig:overview}. Specifically, we design \(D_{\text{xml}}\) in an XML-like format with special tags to represent the hierarchical structure of the document. As shown in Table 1, we design three types of tags to denote the document structure: \mysection, \mysubsection, and \mybr. 
Each section's content is enclosed within \mysection and \mysubsectionend, with a corresponding title that conveys the general meaning of the section. Within each section, there may be several subsections, each enclosed by \mysubsection and \mysubsectionend, with a corresponding subsection title. Additionally, each section and subsection may contain its own content, which is also enclosed within the aforementioned tags. Within the content, we use \mybr to denote paragraph segmentation. With this representation, we convert a structured document tree into a flat textual form. Furthermore, using parsing algorithms, we can easily convert this flat representation back into the original document tree without losing information.

Since most tokens in the flat representation are derived from the original document, we can omit redundant content and restore it during structure recovery.
Therefore, we introduce a \myskip to represent omitted content, retaining only the first and last \(k\) tokens of each paragraph while replacing the omitted content with \myskip. This results in the final \(D_{\text{xml}}\). The parameter \(k\) is a hyperparameter, where a smaller \(k\) reduces the token count but increases parsing errors. 
As shown in Figure 1, the XML-based \(D_{\text{xml}}\) reduces the token count to approximately 1/10 of the original while preserving the original information. We then train a long-context LLM to learn the mapping from the original document \(D\) to \(D_{\text{xml}}\). Based on this learning objective, the model needs to learn how to segment, organize and summarize the total document. Then in the inference stage, we can map predicted \(D_{\text{xml}}\) to \(D_{\text{str}}\) by using the original document and parsing algorithm, obtaining the document tree.

The overall inference process can be describe as follows:
\begin{equation}
    P_{\mathcal{M}}(D_{\text{xml}} \mid D) = \prod_{t=1}^{T} P_{\mathcal{M}}(y_t \mid y_{1:t-1}, D).
\end{equation}
Based on above training objective, the model's generation is a coarse-to-fine process with two steps of iteration:
\paragraph{(1) Structure generation.} The model first generates the hierarchical structure \(S\) (e.g., \(\mysection\) with a suitable title):
\begin{equation*}
P_\mathcal{M}(\mysection \mid D) = \prod_{t=1}^{T_s} P_\mathcal{M}(y_t \mid y_{1:t-1}, D),   
\end{equation*}
where title is automatically generated and fully predicted by the model based on the document.
\paragraph{(2) Content filling based on structure} Then the model generates the content \(C_i\) of each part based on the corresponding \mysection and original document. The content generation probability can be expressed as:
\begin{equation*}
P_\mathcal{M}(C \mid \mysection, D) =  
P_\mathcal{M}(C^{:k}, \myskip, C^{k:}).
\end{equation*}
The model will dynamically skip the middle portion, preserving only the first \(k\) and last \(k\) tokens, and marking the skipped portion with \myskip. 
\paragraph{Wikipedia-Based Label Collection}  
In the previous process, the construction of training label \(D_{\text{xml}}\) requires structural information from the document, which is currently lacking.
Considering that the majority of popular retrieval corpora are derived from Wikipedia, we construct document structure trees based on raw Wikipedia web pages. First, we collect webpage data for Wikipedia entries and remove irrelevant information such as images, links, and references. Then, we extract the entry's core knowledge, obtaining its original  structure such as sections and paragraphs, which is the desired document structure tree \(D_{\text{str}}\). We directly remove the structural information from \(D_{\text{str}}\), retaining only the raw textual content as \(D\). Using the \((D, D_{\text{str}})\) pairs, we can derive the XML-based representation \(D_{\text{xml}}\) following the method described above. Thus, for each webpage, we construct a training dataset containing \((D, D_{\text{xml}})\) pairs, enabling the model to learn how to generate hierarchical document trees from raw plain text.

\subsection{Adaptive Document Refinement}

Based on the structured document tree, we then evaluate each nodes' significance from both local and global perspectives to adapt to varying information requirements and identify the most relevant information.
\paragraph{Local Perspective.} From a local standpoint, relevant information for addressing the query may comprise discrete pieces distributed across multiple paragraphs, which correspond to the leaf nodes of the document tree. As shown in Figure~\ref{fig:overview}, we initiate the refinement process by computing local score(LS) starting at the leaf nodes and subsequently propagating these scores upward through the tree hierarchy.

The scoring mechanism is defined as follows:
\begin{equation*}
    \text{LS}(n_i) =
    \begin{cases}
        \mathcal{M}(\text{query}, n_i) & \text{if } n_i \in \mathcal{N}_L, \\
        \frac{1}{|\mathcal{C}(n_i)|} \sum_{n_j \in \mathcal{C}(n_i)} \text{LS}(n_j) & \text{otherwise}.
    \end{cases}
\end{equation*}

Here, \( M \) represents an universal scoring model used to calculate the similarity between the query and each leaf node, providing the initial local score. \(\mathcal{N}_L\)  represents the set of all leaf nodes in the document tree. These local scores are then propagated to parent nodes by averaging the scores of their child nodes. This method ensures that a parent node's score accurately reflects the overall quality of its content. Importantly, a parent node achieves a high local score only if all its child nodes maintain sufficiently high scores, thereby mitigating the risk of any single child node disproportionately affecting the parent's score.
\paragraph{Global Perspective.} For queries that require comprehensive understanding of the document, it is crucial to evaluate the importance of information from a global perspective. This prevents an overemphasis on localized information points, which could lead to incomplete information retrieval.
The computation of global scores is defined as follows:
\begin{equation*}
    \text{GS}(n_i) =
    \begin{cases}
        \mathbb{I}\left(n_i \in \mathcal{M}(q, \text{outline})\right) & \text{if } n_i \in \mathcal{N}_S, \\
        \dfrac{\text{GS}\left(\text{Pa}(n_i)\right)}{|\mathcal{C}\left(\text{Pa}(n_i)\right)|} & \text{otherwise}.
    \end{cases}
    \label{eq:GS}
\end{equation*}
Here, $\mathbb{I}(\cdot)$ is the indicator function, $\text{Pa}$ represents the parent node， $\mathcal{N}_S$ represents the set of all section nodes in the document tree.
To assess the necessary information from a global standpoint, we fine-tuned the model to select relevant sections based on the query and the document's outline. The document outline consists of the abstract and the titles of all sections. By providing only the outline instead of the entire text, we supply sufficient overall document information while preventing localized details from biasing the model's selection process. From a global perspective, each child node contributes equally to its parent node. Therefore, we uniformly assign each child node to split its parent node's score equally to ensure every child node is considered from global view.
\paragraph{Combination.} We utilize the information scope obtained from the first step as weight to combine each node's local score (LS) and global score (GS), thereby deriving a final measure of each node's contribution for answering the query. The formula is as follows:
\begin{equation*}
    \text{Score}(n_i) = \text{LS}(n_i) + R_q \cdot \text{GS}(n_i).
\end{equation*}
In the final selection process, we sort all nodes based on their node scores and select them sequentially until the designated token budget is met.
To maintain complete structural information in the final refined result, if a parent node is selected, all of its child nodes are automatically included. This ensures the preservation of the document's structural integrity.
Finally, all selected nodes are organized in their original order as they appear in the document and incorporated into the prompt.

\begin{table*}[ht]
\centering
\caption{Overall performance on seven open-domain QA datasets, including single-hop, multi-hop and long-form QA tasks. The best results are in \textbf{bold} and the second are \underline{underlined}. Baselines and our method are limited to 2k token, while full content setting uses complete information without any token limitations and annotated with \textcolor{gray!120}{gray}.}
\label{tab:overall_qa_performance}
\setlength\tabcolsep{4.6pt}
\fontsize{8.9pt}{11pt}\selectfont
\begin{tabular}{p{2.25cm}cccccccccccccc}
\toprule
\multirow{2}{*}{\textbf{Method}} & \multicolumn{2}{c}{\textbf{NQ}} & \multicolumn{2}{c}{\textbf{TriviaQA}} & \multicolumn{2}{c}{\textbf{HotpotQA}} & \multicolumn{2}{c}{\textbf{2Wiki}} & \textbf{ASQA} &  \textbf{ELI5} & \multicolumn{2}{c}{\textbf{PopQA}} & \multirow{2}{*}{\textbf{Tokens}} & \multirow{2}{*}{\textbf{Latency}} \\
\cmidrule(lr){2-3} \cmidrule(lr){4-5} \cmidrule(lr){6-7} \cmidrule(lr){8-9} \cmidrule(lr){10-10}\cmidrule(lr){11-11}\cmidrule(lr){12-13} 
 & Acc & F1 & Acc & F1 & Acc & F1 & Acc & F1 & F1 & F1 & Acc & F1 &  &  \\ 
\midrule
\multicolumn{15}{l}{\textit{\textbf{Vanilla Method}}} \\
Naive Generation & 35.9  & 32.2 & 63.6  & 65.0  & 21.3  & 25.0  & 31.4  & 27.0  & 9.8  & 23.2  & 26.1  & 21.0  & 120  & 1.2  \\ 
Full Content & \textcolor{gray!120}{53.8} & \textcolor{gray!120}{48.1} & \textcolor{gray!120}{70.8} & \textcolor{gray!120}{72.7} & \textcolor{gray!120}{36.0} & \textcolor{gray!120}{42.4} & \textcolor{gray!120}{35.7} & \textcolor{gray!120}{35.7} & \textcolor{gray!120}{34.1} & \textcolor{gray!120}{23.8} & \textcolor{gray!120}{64.1} & \textcolor{gray!120}{49.6} & \textcolor{gray!120}{19567}  & \textcolor{gray!120}{40.6} \\ 
\midrule
\multicolumn{15}{l}{\textit{\textbf{Retrieval-based Method}}} \\
BM25 & 38.1 & 36.0 & 60.3 & 62.2 & 24.7 & 28.9 & 31.8 & 31.5 & 31.1 & 23.7 & 37.5 & 30.6 & 2042  & 3.6 \\ 
Bge-reranker & 40.2 & 37.2 & 59.7 & 62.8 & 25.9 & 31.6 & \underline{33.9} & 27.6 & 30.5 & 23.7 & 37.3 & 29.4 & 2056  & 8.0 \\ 
SBERT & 36.2 & 34.5 & 60.0 & 62.1 & 25.6 & 29.7 & 33.2 & 26.4 & 30.3 & \underline{23.8} & 38.8 & 31.1 & 2054  & 7.0 \\ 
Recomp-ext & 38.0 & 35.0 & 59.4 & 61.3 & 25.7 & 30.3 & 30.4 & 26.2 & 30.8 & \underline{23.8} & 36.2 & 28.9 & 1915  & 7.2 \\ 
\midrule
\multicolumn{15}{l}{\textit{\textbf{Semantic Chunking Method}}} \\
Jina-Segment & 40.0 & 38.3 & 61.3 & 63.7 & 26.0 & 31.3 & 32.4 & 26.4 & 31.2 & 23.7 & 36.3 & 28.8 & 2148  & 8.4 \\ 
Meta-Chunking & 39.0 & 37.4 & 61.7 & 63.8 & 26.7 & 31.7 & 33.1 & 27.3 & 30.7 & 23.7 & 35.6 & 28.8 & 2181  & 8.6 \\ 
\midrule
\multicolumn{15}{l}{\textit{\textbf{Perplexity-based Methods}}} \\
Selective-Context & 36.1 & 35.0 & 64.4 & 67.5 & 24.0 & 29.5 & 28.8 & 25.2 & 28.6 & 23.1 & 45.3 & 40.2 & 1841  & 100.6 \\ 
LLMLingua2 & 44.4 & \underline{43.0} & 66.9 & \underline{69.8} & 28.3 & 36.9 & 29.4 & 32.2 & 29.9 & 23.4 & 51.1 & 39.9 & 2043  & 21.6 \\ 
LongLLMLingua & \underline{45.4} & 42.4 & \underline{67.6} & \underline{69.8} & \underline{34.7} & \underline{41.7} & 33.1 & \underline{34.5} & \underline{33.6} & 23.7 & \underline{56.8} & \underline{43.6} & 1976  & 496.6 \\ 
\midrule
\multicolumn{15}{l}{\textit{\textbf{Hierarchical Method}}} \\
\rowcolor[RGB]{236,244,252} 
LongRefiner(Ours) & \textbf{54.4} & \textbf{48.9} & \textbf{71.7} & \textbf{73.0} & \textbf{39.3} & \textbf{45.8} & \textbf{36.1} & \textbf{35.0} & \textbf{35.8} & \textbf{23.9} & \textbf{59.9} & \textbf{45.9} & 1933  & 10.8 \\ 
\bottomrule
\end{tabular}
\end{table*}

\subsection{Training and Inference}
In order to improve the usability and efficiency of our method, we have made the following designs during the training and inference processes.
\paragraph{Training.} Our approach involves three training tasks: query analysis, document structuring, and global selection, all trained on a single base model. Due to significant variations in input token length of these tasks, mixed training will introduce excessive padding, reducing efficiency. To address this challenge, we employ task-specific LoRA modules for each task, with each module's parameters accounting for only 0.03\% of the total model parameters. This design enables \ours{} to share the same backbone while switching between different tasks through plug-and-play task-specific parameter loading. Notably, this approach maintains inference latency comparable to a shared module while preventing task interference.
\paragraph{Inference.} To reduce latency, inference is split into offline and online stages. 
In offline stage, the model will perform hierarchical document structuring task for the documents in corpus. And in online stage, it performs analysis of the user's query followed by adaptive refinement to generate the final output. Since the online stage only involves processing hundreds of input tokens and generating dozens of output tokens, the overall latency is only about 25\% of the standard setting.

\section{Experimental Settings}
\subsection{Datasets and Evaluation Metrics}
We conduct experiments on seven widely used datasets in three types: Single-hop QA (NQ~\cite{nq}, TriviaQA~\cite{triviaqa}, PopQA~\cite{popqa}), Multi-hop QA(HotpotQA~\cite{hotpotqa}, 2WikiMultiHopQA~\cite{2wiki}), and Long-form QA(ASQA~\cite{asqa}, ELI5~\cite{eli5}). We use Accuracy and F1 Score as metrics for evaluation. Detailed information is provided in Appendix~\ref{appendix:implementation_details}.


\subsection{Baselines}

We compare our approach with three categories of baseline methods: (1) Retrieval-based Methods: BM25~\cite{bm25}, Bge-Reranker~\cite{bge}, SBERT~\cite{sentencebert}, and Recomp~\cite{recomp}; (2) Semantic Chunking Methods: Jina Segmenter (accessed via API) and Meta-Chunking~\cite{metachunking}; (3) Perplexity-based Methods: Selective-Context~\cite{selective-context}, LongLLMLingua~\cite{longllmlingua}, and LLMLingua2~\cite{llmlingua2}. Detailed descriptions are provided in Appendix~\ref{appendix:implementation_details}.

\subsection{Implementation Details}

We use Llama3.1-8B-Instruct~\cite{llama3} as the generator with 64k context window size to accommodate all documents. We construct the corpus based on the full Wikipedia 2018 dump~\cite{dpr} and follow the MaxP~\cite{maxp_deeper_text_understanding} design in LongRAG~\cite{jiang2024longrag} to retrieve the top-8 full documents for each query. Our refiner is based on the Qwen 2.5-3B-Instruct~\cite{qwen2.5}, and we train the model using the LoRA method. For additional details, please refer to Appendix~\ref{appendix:implementation_details}.

\subsection{Experimental Results}
\paragraph{Overall Performance.} As shown in Table~\ref{tab:overall_qa_performance}, we evaluate our method against various refinement approaches across seven diverse datasets under a fixed constraint of 2k tokens. We have several findings:
(1) Our method achieves \textbf{the best performance across all datasets while maintaining low latency}, demonstrating the effectiveness of leveraging internal document structure for refinement. The consistent improvements across different query types validate the efficacy of our adaptive design.
(2) While existing methods excel in either performance or latency, our approach maintains latency comparable to retrieval-based approaches while surpassing the performance of perplexity-based methods by more than 9\%.
(3) Compared to vanilla approach using complete documents, our method demonstrates remarkable efficiency by \textbf{achieving superior performance on six datasets while reducing token usage by 10x and latency by 4x}. 
The exception is PopQA, where documents are relatively short with minimal noise, enabling effective LLM comprehension of complete documents. Our method's potential information loss may slightly impact performance in such low-noise scenarios.
\paragraph{Ablation Study.} 
\begin{table}[!t]
\centering
\caption{Ablation study on three types of datasets.}
\label{tab:ablation_study}
\setlength\tabcolsep{4pt}
\setlength\tabcolsep{3pt} 
\fontsize{7pt}{9pt}\selectfont
\begin{tabular}{lccccc}
\toprule
\multirow{2}[2]{*}{\textbf{Method}} & \textbf{Single-hop} & \textbf{Multi-hop} & \textbf{Long-form} \\
 & (EM) & (Acc) & (F1) \\
\midrule
\rowcolor[RGB]{236,244,252} 
LongRefiner & \textbf{62.3} & \textbf{37.4} & \textbf{30.2} \\ 
\midrule
w/o Query Analysis & 60.3 & 36.2 & 29.6 \\
w/o Doc. Structuring & 45.7 & 29.9 & 27.1 \\
w/o Adaptive Refine. & 57.7  & 35.3 & 29.2 \\ 
\bottomrule
\end{tabular}
\end{table}

\begin{figure}[!t]
\centering
\includegraphics[width=0.9\linewidth]{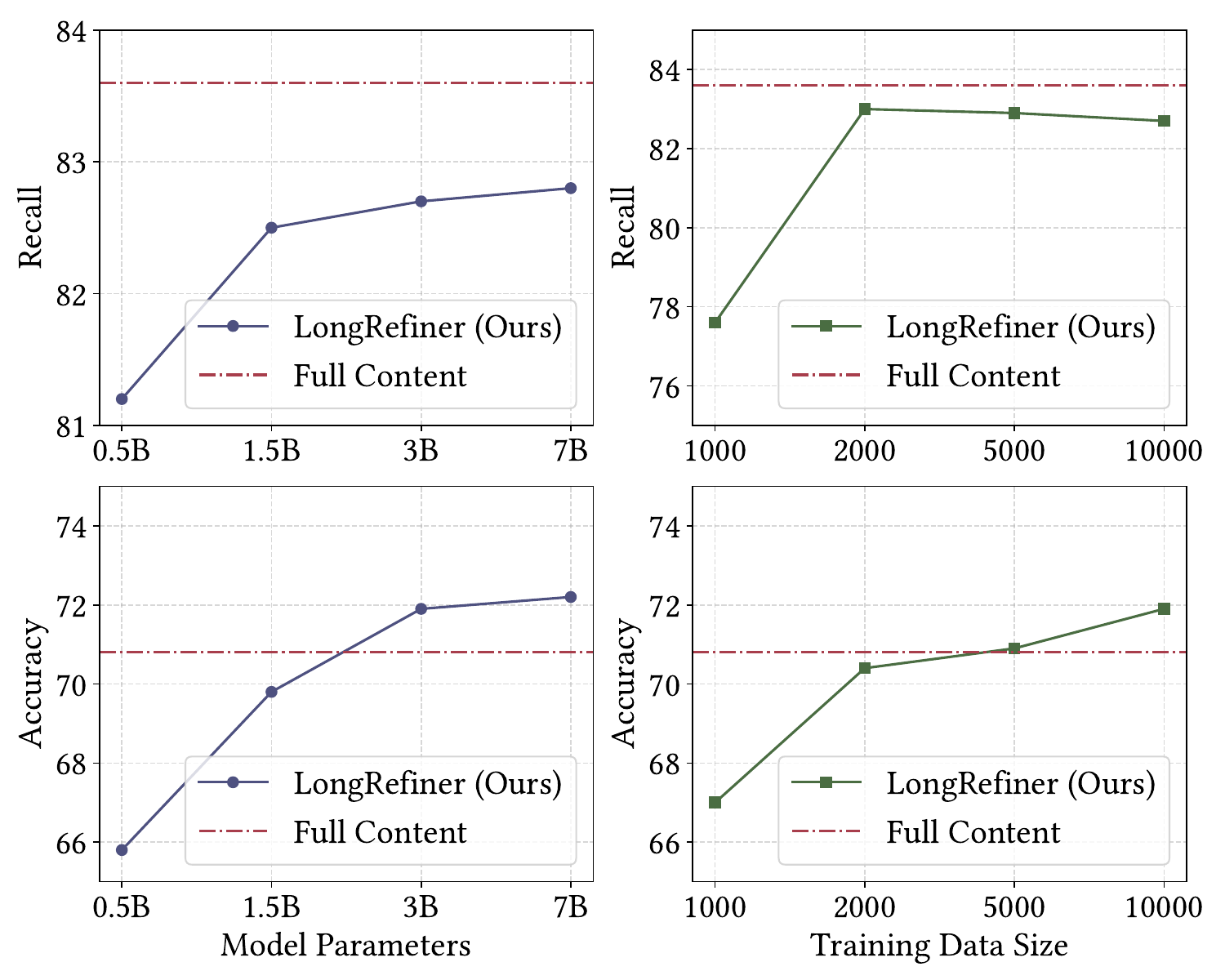}
\caption{Analysis of scaling the base model size (left) and training data amount (right) in hierarchical document structuring step on TriviaQA. Recall represents the proportion of golden answer in the input prompt.}
\label{fig:step2_experiments}
\end{figure}

\begin{figure}[!t]
\centering
\includegraphics[width=0.8\linewidth]{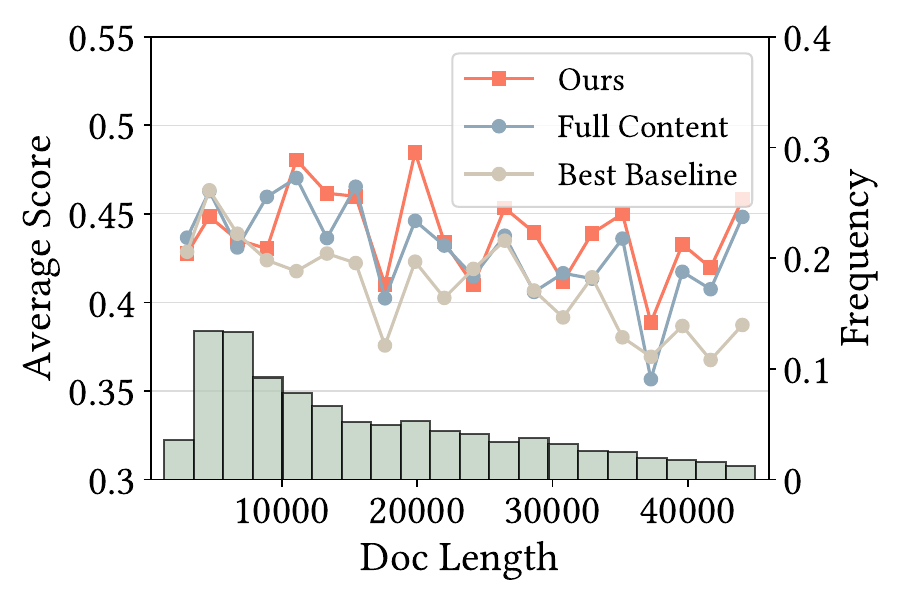}
\caption{The performance of \ours{} across different document lengths, where Doc Length refers to the total number of tokens for all retrieved documents corresponding to a query.}
\label{fig:doc_length_performance}
\end{figure}

To quantify the contributions of different components in our framework, we conduct ablation studies on the three key modules, yielding results in Table ~\ref{tab:ablation_study}. The findings can be summarized as follows:
(1) Removing any step results in significant performance degradation across all query types, demonstrating the necessity and effectiveness of all three components in our system.
(2) The Hierarchical Structuring module proves most crucial, with its removal resulting in nearly 20\% degradation.
This substantial impact stems from its fundamental role in modeling document structure, without which the method degrades to basic chunking. 
(3) The performance decrements from removing either query analysis or adaptive refinement modules confirm the effectiveness of our dual-perspective approach in evaluating information quality from both global and local viewpoints.
\paragraph{Scaling Analysis of Document Structuring Model.}
Accurate hierarchical document structuring is crucial for refinement quality. To evaluate its practical applicability, we analyze our model's scalability by evaluating the effects of model size and training data volume on performance.

\textbf{Parameter Aspect.} We finetune models of varying sizes from Qwen series and measure both refinement recall and answer accuracy. 
As shown in Figure~\ref{fig:step2_experiments}, performance improves with increased model parameters, though with diminishing returns. The experiments demonstrate that with sufficient training data, larger models can better capture document structure, approaching full-content baseline recall rates while achieving superior QA performance due to reduced noise. 

\textbf{Data Aspect.} Training data scaling reveals an intriguing pattern: while QA performance consistently improves, recall shows initial increase followed by slight decline.  Case analysis shows that the temporary recall increase stems from underfitting. With limited data, the model develops weaker structuring capabilities, resulting in fewer sections and larger content blocks. Although this approach mirrors the structure of the original document more closely, the structural information becomes less accurate and comprehensive, which hampers subsequent refinement. However, with sufficient training data, models can generate more authentic and complete document structures. In this case, although some recall loss occurs due to XML-format parsing errors, the improved structural accuracy compensates for this loss.
\paragraph{Performance on Different Document Lengths.}
We evaluate our method's effectiveness across varying document lengths, comparing it with full content setting and best baseline. As shown in Figure \ref{fig:doc_length_performance}, we have two key findings: 
(1) While our method shows relatively lower performance on shorter documents compared to full-document methods, it significantly outperforms them on longer texts. This can be attributed to the lower noise levels in shorter documents, which minimally impact model generation. Notably, our method achieves this with 10x fewer tokens than the full-content approach, demonstrating its viability even for shorter texts. 
(2) Our method consistently and significantly outperforms LongLLMLingua across almost all document lengths, highlighting the substantial advantages of leveraging structural information for text refinement over perplexity-based approaches.

\begin{table}[!t]
\centering
\caption{Analysis of using different model to calculate local scores.}
\label{tab:scoring_model_impact}
\setlength\tabcolsep{4pt}
\fontsize{8.6pt}{10.4pt}\selectfont
\begin{tabular}{lccccc}
\toprule
\multirow{2}[2]{*}{\textbf{Scoring Model}} & \multicolumn{2}{c}{\textbf{Single-hop}} & \multicolumn{2}{c}{\textbf{Multi-hop}} & \textbf{Long-form} \\
\cmidrule(lr){2-3} \cmidrule(lr){4-5} \cmidrule(lr){6-6}
 & Acc & F1 & Acc & F1 & F1  \\
\midrule
BM25 & 52.7 & 49.5 & 34.9 & 37.6 & 28.9 \\ 
E5 & 60.8 & 55.0 & 36.4 & 38.7 & 29.5 \\ 
SBERT & 58.6 & 53.1 & 36.1 & 38.4 & 29.5 \\ 
Bce-reranker & 60.8 & 54.8 & 36.6 & 40.0 & 29.5 \\ 
Bge-reranker & 62.0 & 55.9 & 37.7 & 40.4 & 29.9 \\ 
\midrule
\rowcolor[RGB]{229, 231, 233 }
Best Baseline & 56.6 &  51.9 & 33.9 & 38.1 & 28.7 \\
\bottomrule
\end{tabular}
\end{table}
\paragraph{Impact of Different Scoring Model}
We evaluate the impact of various scoring models for computing local score, including term-based method, embedding models, and rerankers. As shown in Table~\ref{tab:scoring_model_impact}, all methods except term-based approaches outperform the best baseline. The reranker model achieves the best performance by accurately capturing relevance between local paragraphs and queries. While embedding-based scoring had slightly lower performance, it offers superior computational efficiency, making it a practical alternative. Notably, the choice of scoring model has the greatest impact on single-hop datasets, where local information are critical to overall performance.

\section{Related Works}
\paragraph{Retrieval-augmented Generation.}
Retrieval-augmented Generation (RAG)~\cite{rag,retro,realm} enhances LLMs by incorporating retrieved knowledge into input prompts, reducing hallucination and knowledge limitations. While retrieving multiple documents ensures high recall, it introduces challenges: lengthy documents contain noise and irrelevant information, increasing both computational costs and potential output errors. Current solutions either focus on improving LLMs' long-text processing capabilities~\cite{longalign,longlora} or refining retrieved knowledge~\cite{recomp} for flexible deployment across different models.
\paragraph{Knowledge Refinement Methods.}
Knowledge refinement methods can be categorized into two approaches~\cite{promptcompress_survey}: (1) Hard Prompt Refinement, which directly processes text through token removal~\cite{selective-context,longllmlingua,llmlingua,llmlingua2}, summarization~\cite{bider_jiajie,recomp,prca,24_acl_chunking_free}, or chunk-based selection~\cite{compact,dparag,filco_zhiruowang}. This approach requires no model adaptation and offers better interpretability. (2) Soft Prompt Refinement, which encodes documents into vector or semantic spaces~\cite{xrag_compression,lloco_compress,flexrag}, but requires additional training. While existing methods struggle with long texts or lack comprehensive document understanding, our method addresses these limitations through structured document modeling.

\section{Conclusion}

In this paper, we presented \ours{}, a document-level refinement framework that effectively addresses the challenges of processing long retrieved documents in RAG systems. By integrating dual-level query analysis, hierarchical document structuring, and adaptive refinement capabilities through multi-task LoRA learning, our approach significantly improves both the efficiency and accuracy of long document refinement. Experiments show that \ours{} consistently outperforms existing baselines, achieving superior generation quality while having lower latency. These results validate the effectiveness of our document-level approach in leveraging hierarchical textual information for efficient RAG systems. 

\section{Limitations}

Although \ours{} demonstrates strong performance and low latency across various datasets, there remain several limitations that warrant further exploration and improvement. First, enhancing support for diverse data types: In real-world scenarios, retrieved documents often contain not only plain text but also tables, images, and hyperlinks. How to model and refine such content with complex information structures remains an unsolved challenge. This may involve extending our XML-based syntax to accommodate these varied data types and training a more versatile refinement model using real-world data. Second, our current approach relies entirely on general-domain Wikipedia corpus, making it challenging to directly transfer to vertical domains such as enterprise or finance, where document characteristics may differ significantly. In such scenarios, we may need to design and model specifically for their documents and use cases, potentially leveraging teacher LLMs for text structure annotation. This represents another important direction for future exploration.

\bibliography{reference}

\clearpage
\appendix
\section*{Appendix}


\section{Implementation Details}
\label{appendix:implementation_details}

\subsection{Dataset and Evaluation Metrics}

To comprehensively evaluate the performance of our method across different query types, we selected seven widely used datasets categorized into three types: (1) Single-hop QA: NQ~\cite{nq}, TriviaQA~\cite{triviaqa}, and PopQA~\cite{popqa}; (2) Multi-hop QA: HotpotQA~\cite{hotpotqa} and 2WikiMultiHopQA~\cite{2wiki}; (3) Long-form QA: ASQA~\cite{asqa} and ELI5~\cite{eli5}. Notably, the PopQA dataset does not include a training set and is therefore utilized as out-of-domain data to assess the generalization capability of our approach.
For each question, we follow the retrieval approach in LongRAG~\cite{jiang2024longrag} to retrieve top-8 full documents from Wikipedia Dump 2018. 
For the five short-answer QA datasets, we used Accuracy and F1 Score as the evaluation metrics for five short-answer datasets and use F1 Score for two long-form datasets. Additionally, we recorded the number of generator input tokens  and the overall online latency to evaluate the efficiency of each method.

\subsection{Baseline Details}

Our baseline approaches encompass various techniques for refining and compressing retrieved results in long-text scenarios, categorized into three types:
(1) Retrieval-based: These methods segment the retrieved long documents into fixed-length chunks and utilize a scoring model to select the chunks with the highest similarity to the query. We employed four scoring models: BM25~\cite{bm25}, Bge-Reranker~\cite{bge}, SBERT~\cite{sentencebert}, and Recomp~\cite{recomp}. (2) Semantic Chunking : These techniques divide long texts into semantically coherent chunks and then select relevant chunks based on similarity. We used two methods for this purpose: Jina Segmenter (accessed via API) and Meta-Chunking~\cite{metachunking}; (3) Perplexity-based: These methods focus on refining long texts by leveraging perplexity measures. The specific methods implemented include Selective-Context~\cite{selective-context}, LongLLMLingua~\cite{longllmlingua}, and LLMLingua2~\cite{llmlingua2}.

For all baselines, we employ Llama3.1-8B-Instruct~\cite{llama3} as the generator model. Both refinement and compression processes are consistently performed on the top-8 retrieved long documents for each query to ensure consistency and fairness. Additionally, the prompts used for the generator remain same across all baselines and our proposed method to ensure fairness in answer generation.

\paragraph{Retrieval-based Methods.}
For retrieval-based methods, we first segment each long document into chunks of six sentences using the nltk library~\cite{nltk}. Sentence-based segmentation is chosen to preserve intra-chunk coherence, resulting in chunks of approximately 200 tokens each, which balances effectiveness and fairness. Subsequently, all chunks are ranked using the bge-reranker model, which achieves the best performance in our experiments. Chunks are selected sequentially based on a token budget. When incorporating a chunk, we also include its corresponding document title to enhance the model’s understanding of the context.

\paragraph{Semantic Chunking Methods.}
We employ two semantic chunking methods, which differ from retrieval-based methods in that they use automated, intelligent chunking instead of manually defined granularities. For the Jina-Segment method, we use the segmentation API provided by Jina-AI, setting the maximum chunk length to 500 tokens to avoid overly large chunks, with all other API parameters left at default values. For the Meta-Chunking method~\cite{metachunking}, we use the official implementation with the ppl method and set the threshold to 0.5.

\begin{table*}[ht]
\centering
\caption{Overall performance on seven open-domain QA datasets, including single-hop, multi-hop and long-form QA tasks. The best results are in \textbf{bold} and the second are \underline{underlined}. Baselines and our method are limited to 2k token, while full content uses complete information and annotated with \textcolor{gray!120}{gray}. The generator is Qwen2.5-7B-Instruct.}
\label{tab:change_generator}
\setlength\tabcolsep{4.6pt}
\fontsize{8.9pt}{11pt}\selectfont
\begin{tabular}{p{2.25cm}cccccccccccccc}
\toprule
\multirow{2}{*}{\textbf{Method}} & \multicolumn{2}{c}{\textbf{NQ}} & \multicolumn{2}{c}{\textbf{TriviaQA}} & \multicolumn{2}{c}{\textbf{HotpotQA}} & \multicolumn{2}{c}{\textbf{2Wiki}} & \textbf{ASQA} &  \textbf{ELI5} & \multicolumn{2}{c}{\textbf{PopQA}} & \multirow{2}{*}{\textbf{Tokens}} & \multirow{2}{*}{\textbf{Latency}} \\
\cmidrule(lr){2-3} \cmidrule(lr){4-5} \cmidrule(lr){6-7} \cmidrule(lr){8-9} \cmidrule(lr){10-10}\cmidrule(lr){11-11}\cmidrule(lr){12-13} 
 & Acc & F1 & Acc & F1 & Acc & F1 & Acc & F1 & F1 & F1 & Acc & F1 &  &  \\ 
\midrule
\multicolumn{15}{l}{\textit{\textbf{Vanilla Method}}} \\
Naive Generation & 24.5 & 23.5 & 47.2 & 49.0 & 21.2 & 28.0 & 26.6 & 32.0 & 26.0 & 18.2 & 18.2 & 13.0 & 120.1 & 1.2 \\ 
Full Content & \textcolor{gray!120}{47.6} & \textcolor{gray!120}{32.5} & \textcolor{gray!120}{65.6} & \textcolor{gray!120}{53.2} & \textcolor{gray!120}{37.5} & \textcolor{gray!120}{31.6} & \textcolor{gray!120}{36.5} & \textcolor{gray!120}{27.9} & \textcolor{gray!120}{26.5} & \textcolor{gray!120}{24.2} & \textcolor{gray!120}{59.8} & \textcolor{gray!120}{35.3} & \textcolor{gray!120}{19566.6} & \textcolor{gray!120}{40.6} \\ 
\midrule
\multicolumn{15}{l}{\textit{\textbf{Retrieval-based Method}}} \\
BM25 & 31.9 & 25.9 & 53.8 & 53.2 & 23.3 & 27.9 & 29.3 & 30.1 & 31.6 & 25.0 & 34.2 & 21.0 & 2042.0 & 3.6 \\ 
Bge-reranker & 36.5 & 29.7 & 53.6 & 51.7 & 24.5 & 28.9 & 28.2 & 28.3 & 30.3 & 24.8 & 35.0 & 22.0 &  2055.6 & 8.0 \\ 
SBERT & 33.6 & 26.9 & 52.4 & 51.9 & 24.8 & 27.7 & \underline{31.0} & 29.4 & 30.3 & 24.6 & 36.5 & 22.9 & 2054.3 & 7.0 \\ 
Recomp-ext & 31.6 & 25.4 & 54.4 & 53.1 & 24.2 & 28.6 & 27.2 & 27.4 & 31.0 & 24.9 & 31.6 & 20.8 & 1914.6 & 7.2 \\ 
\midrule
\multicolumn{15}{l}{\textit{\textbf{Semantic Chunking Method}}} \\
Jina-Segment & 35.1 & 28.4 & 55.8 & 54.3 & 25.6 & 30.3 & 28.4 & 30.2 & 30.8 & 24.8 & 29.5 & 17.8 & 2148.0 & 8.4 \\ 
Meta-Chunking & 33.6 & 27.8 & 52.7 & 52.6 & 24.3 & 27.8 & 28.1 & 27.2 & 31.0 & 24.8 & 32.2 & 20.2 & 2180.6 & 8.6 \\ 
\midrule
\multicolumn{15}{l}{\textit{\textbf{Perplexity-based Methods}}} \\
Selective-Context & 22.2 & 20.5 & 45.4 & 48.2 & 21.0 & 26.4 & 25.5 & 28.0 & 26.5 & 24.3 & 37.4 & 24.6 & 1841.0 & 100.6 \\ 
LLMLingua2 & 33.4 & 29.9 & 54.3 & 55.6 & 25.8 & 33.4 & 28.4 & 30.8 & 32.5 & 24.6 & 43.6 & 26.5 & 2043.0 & 21.6 \\ 
LongLLMLingua & \underline{40.4} & \underline{33.9} & \underline{60.1} & \underline{60.0} & \underline{32.7} & \underline{38.2} & 30.8 & \underline{32.9} & \underline{34.6} & \underline{25.0} & \underline{55.1} & \underline{33.9} & 1976.4 & 496.6 \\ 
\midrule
\multicolumn{15}{l}{\textit{\textbf{Hierarchical Method}}} \\
\rowcolor[RGB]{236,244,252} 
LongRefiner(Ours) & \textbf{50.6} & \textbf{40.6} & \textbf{69.4} & \textbf{67.7} & \textbf{38.3} & \textbf{42.7} & \textbf{33.2} & \textbf{33.4} & \textbf{38.5} & \textbf{25.2} & \textbf{58.4} & \textbf{36.2} & 1933.0 & 10.8 \\ 
\bottomrule
\end{tabular}
\end{table*}

\paragraph{Perplexity-based Methods.}
These methods are implemented using the FlashRAG framework~\cite{flashrag_jiajie}, closely following the official implementation. For the Selective-Context method, we set the compression granularity to the token level. For the other two methods, we adhere to their default configurations. We adjust the compression ratio and target token count in all three methods to ensure the final number of tokens remains within the token budget.

\subsection{Training Details}

\paragraph{Training Setup.}
The training process leverages Llama-Factory\cite{llamafactory} with LoRA fine-tuning. The base model used is Qwen2.5-3B-Instruct\cite{qwen2.5}. The three steps in our method utilize maximum sequence lengths of 2k, 32k, and 4k, respectively. The per-device batch size is set to 1, gradient accumulation to 8, learning rate to $3e^{-5}$, and the warmup ratio to 0.1, with bf16 precision enabled. Each task is trained for 1 epoch on 4 NVIDIA A800 GPUs.

\paragraph{Training Data.}
The training dataset is constructed using the version collected by FlashRAG. We use the first 10,000 samples from the training set of each dataset, which are merged to form the final dataset. For the Dual-Level Query Analysis and Adaptive Document Refinement tasks, training labels are generated using Llama3.1-70B-Instruct, with instructions provided in the appendix. For the Hierarchical Document Structuring task, we preprocess the Wikipedia dump provided by KILT~\cite{kilt}. Non-relevant information (e.g., references, external links) is removed, and scripts are written to extract structural information and text. These scripts are available in our code repository.

\subsection{Inference Details}

Both \ours and the generator inference are implemented using the VLLM framework~\cite{vllm}, with the temperature set to 0 for greedy decoding to eliminate randomness in results. The maximum number of output tokens is set to 500 to avoid truncation. In our method, local score computation utilizes the bge-reranker-v2-m3~\cite{bge}. To ensure fairness, we use the same prompts as the baselines, modifying only the refinement results accordingly. When structuring the input, we account for hierarchical relationships: if all child chunks of a parent chunk are selected, the parent chunk is also selected to ensure completeness. Similarly, if a parent chunk is selected, all its child chunks are included in the input.

\section{Impact of Different Base Generator}

To verify the robustness of our approach across different generators, we conducted additional experiments using Qwen2.5-7B-Instruct, keeping all other settings consistent with the main results. As shown in Table~\ref{tab:change_generator}, our method also demonstrated strong performance on Qwen2.5-7B-Instruct, significantly outperforming all baseline methods across all datasets.

\section{Case Study}

We present a retrieved long document and its XML-based refinement result. As shown in Table~\ref{table:case_study}, the left column displays the original plain text, while the right column shows the model’s output, with each section condensed into a single line for clarity. The original document lacks structural information, making it difficult to quickly grasp its content or locate key information. In contrast, the refined result uses XML-based tags to reveal the document’s structure and employs \myskip to omit redundant details, resulting in a more concise output. Using our designed syntax, we can efficiently parse specific document fragments from the model’s output through regular expression matching.

\begin{table*}[!tbp]
\centering
\caption{An example about the original document and the results of hierarchical modeling.}
\label{table:case_study}
\begin{tabular}{|p{0.7\linewidth}|p{0.3\linewidth}|}
\hline
\textbf{Document (Full Content)} & \textbf{Refined Results} \\
\hline
Bunk'd is an American comedy television series created by Pamela Eells O'Connell that premiered on Disney Channel on July 31, 2015, and is a spinoff of ``Jessie''. The series stars Peyton List, Karan Brar, and Skai Jackson from ``Jessie'', as well as Miranda May.  & 
\textbf{\textcolor{myblue}{<abstract>}} Bunk'd is an \textbf{\textcolor{myred}{<skip>}}  as well as Miranda May.\textbf{\textcolor{myblue}{</abstract>}} \\
\hline
Emma, Ravi, and Zuri leave New York City and head off to Moose Rump, Maine, to spend the summer at Camp Kikiwaka, where their parents met as teenagers. The Ross children and their new friends try their best to adapt to their lives at the camp, which was founded by Jedediah Swearengen and is named after a legendary creature that lives in the nearby forest. %
In ``We Didn't Start the Fire'', several cabins at Camp Kikiwaka are destroyed by a fire after a candle was left unattended. In the premiere of the third season, ``We Can't Bear It'', the Ross children return with a new generation of campers to find the cabins were never rebuilt and Gladys ran away with the insurance money. The Ross children then convince their parents to buy Camp Kikiwaka and put them in charge. 
& 
\textbf{\textcolor{myblue}{<section: Plot>}}
Emma, Ravi, and Zuri \textbf{\textcolor{myred}{<skip>}} creature that lives in the nearby forest.\textbf{\textcolor{myred} {<br>}} In ``We Didn't Start the Fire'' \textbf{\textcolor{myred}{<skip>}} waka and put them in charge. 
\textbf{\textcolor{myblue}{</section: Plot>}} \\
\hline
 A few cast members from ``Jessie'' reprise their roles in ``Bunk'd''. Cameron Boyce appears as a special guest star in ``Luke's Back'' and ``Luke Out Below'', reprising his role as Luke Ross. Kevin Chamberlin appears as a special guest star in ``A Bad Case of the Weasels'', reprising his role as Bertram, the butler. Christina Moore appears as a guest star in ``Mother May I?'' reprising her role of Christina Ross. & 
\textbf{\textcolor{myblue}{<section: Cast>}}\textbf{\textcolor{myblue}{<sub-section: Main cast>}}
A few cast members from ``Jessie'' \textbf{\textcolor{myred}{<skip>}} reprising her role of Christina Ross.
\textbf{\textcolor{myblue}{</sub-section: Main cast>}}\textbf{\textcolor{myblue}{</section: Cast>}} \\
\hline
The series is a spinoff of ``Jessie''. ``Bunk'd'' was renewed for a second season by Disney Channel on February 29, 2016. The second season premiered on August 23, 2016. 
The series was renewed for a third season by Disney Channel on August 31, 2017. On June 1, 2018, it was announced that Peyton List, Karan Brar, Skai Jackson, and Miranda May would be returning for the third season and that Raphael Alejandro, Will Buie Jr., and Mallory Mahoney would be joining the cast. The third season premiered on Disney Channel on June 18, 2018. 
In March 2018, actress Skai Jackson stated in an interview that she was leaving Disney and that \"Bunk'd\" would end with the third season. In September 2018, it was confirmed in a report from \"The Hollywood Reporter\" that Peyton List would also leave the series after the conclusion of its third season.  
On November 15, 2018, it was announced by Disney Channel that the series was renewed for a fourth season. Miranda May, Mallory James Mahoney, Raphael Alejandro, and Will Buie Jr. will be returning for the fourth season, with new unannounced cast also set to star alongside them. Peyton List, Karan Brar, and Skai Jackson will not be returning for the fourth season. Additionally, \"Andi Mack\"s Phil Baker and Erin Dunlap will take over as executive producers in the fourth season. Production for the fourth season is scheduled to begin in March 2019. & 
\textbf{\textcolor{myblue}{<section: Production>}}\textbf{\textcolor{myblue}{<sub-section: Season 2>}}
``Bunk'd'' was renewed for a \textbf{\textcolor{myred}{<skip>}} 3, 2016.
\textbf{\textcolor{myblue}{</sub-section: Season 2>}}
\textbf{\textcolor{myblue}{<sub-section: Season 3>}}
The series was renewed for a third season \textbf{\textcolor{myred}{<skip>}} 8, 2018.
\textbf{\textcolor{myblue}{</sub-section: Season 3>}}
\textbf{\textcolor{myblue}{<sub-section: Season 4>}}
In March 2018 \textbf{\textcolor{myred}{<skip>}} after the conclusion of its third season. \textbf{\textcolor{myred}{<br>}} On November 15, 2 \textbf{\textcolor{myred}{<skip>}} in March 2019.\textbf{\textcolor{myblue}{</sub-section: Season 4>}}
\textbf{\textcolor{myblue}{</section: Production>}}
 \\
\hline
In Canada, the series premiered on Disney Channel Canada on the second day of the channel's launch on September 2, 2015. The series premiered on Disney Channels in the United Kingdom and Ireland on November 20, 2015, and premiered in Australia and New Zealand on January 14, 2016. &
\textbf{\textcolor{myblue}{<section: Broadcast>}}
In Canada, the series premiered on Disney \textbf{\textcolor{myred}{<skip>}} 4, 2016.
\textbf{\textcolor{myblue}{</section: Broadcast>}}\\
\hline
\end{tabular}
\end{table*}

\section{Instructions}
\label{appendix:instructions}

\paragraph{Annotation Instruction}
In both the dual-level query analysis and adaptive document refinement stages, we annotate training labels using Llama3.1-70B-Instruct. The prompts used for annotation are provided in Prompt A and Prompt B. For global search annotation, the document abstracts and outlines are generated by our trained model rather than using golden data. This approach ensures a closer simulation of real-world inference scenarios.

\paragraph{Generation Instruction}
Our method and all baselines employ the same generation prompt to ensure fairness. The specific prompts are detailed in Prompt C.1 and Prompt C.2. For short-form datasets, we instruct the model to output responses with a fixed prefix (e.g. So the final answer is), and the corresponding short answers are extracted using regular expressions for evaluation. For long-form datasets, where responses are inherently more extensive, the model's generated outputs are directly used for final evaluation without additional processing.

\onecolumn
\begin{tcolorbox}[title={Prompt A: Annotate the required information types for the query},label=gpt] 
\label{prompt_annot_query}
You are an assistant that performs step-by-step analysis of user queries.\newline

**Instructions for Query Analysis:**

When given a query, please **understand the query intents**, and classify the query as either **[Local]** or **[Global]**.\newline
- **[Global]**: The query requires a broad or vague range of knowledge (e.g., summary or open-ended questions), and may require a comprehensive understanding of the document.

- **[Local]**: The query has a clear and fixed answer with a narrow scope of knowledge (e.g., factual questions), and only a small amount of text fragments are needed to answer.\newline

**Output Format:**

Please present the results in JSON format with the following keys:\newline
**query\_type**: [Local] or [Global]\newline

**Demonstration**\newline
\{\textit{demonstrations}\}\newline

Query: \{\textit{query}\}\newline
Results:\newline
\end{tcolorbox}

\begin{tcolorbox}[title={Prompt B: Annotate the selected titles in global search},label=gpt] 
\label{prompt_select_title}
You will be provided with three inputs:\newline
1. A question.\newline
2. The abstract of a document.\newline
3. Outline of the document, contains titles of section and subsections in the document.\newline

Your task is to understand the article based on its abstract and outline, and select all the parts that are helpful for answering questions (provide corresponding titles, or `abstract`).\newline

**Demonstration**\newline
\{\textit{demonstrations}\}\newline

Document abstract: \{\textit{abstract}\}\newline
**Document outline**: \{\textit{outline}\}\newline
**Question**:\{\textit{question}\}\newline
Output:\newline
\end{tcolorbox}

\begin{tcolorbox}[title={Prompt C.1: Prompt for generator to provide answer (for short-form dataset)},label=gpt] 
\label{prompt_c_1}
Find the useful content from the provided documents, then answer the question. Answer the question directly. Your response should be very concise. Please provide use 'So the final answer is:' as a prefix for the final answer.\newline

Output format:\newline
\{\textit{demonstrations}\}\newline

The following are given documents.\newline
\{\textit{reference}\}\newline

Answer the question directly. Your response should be very concise. Please provide use 'So the final answer is:' as a prefix for the final answer.\newline

**Question**: \{\textit{question}\}\newline
**Response**:\newline
\end{tcolorbox}

\begin{tcolorbox}[title={Prompt C.2: Prompt for generator to provide answer (for long-form dataset)},label=gpt] 
\label{prompt_c_2}
Find the useful content from the provided documents, then answer the question. Answer the question directly. Your response should be very detailed. \newline

Output format:\newline
\{\textit{demonstrations}\}\newline

The following are given documents.\newline
\{\textit{reference}\}\newline

Answer the question directly. Your response should be very detailed.\newline

**Question**: \{\textit{question}\}\newline
**Response**:\newline
\end{tcolorbox}

\twocolumn

\end{CJK}
\end{document}